\documentclass[runningheads]{llncs}

\usepackage{eccv}

\usepackage{eccvabbrv}

\usepackage{graphicx}
\usepackage{booktabs}
\usepackage{multirow} 
\usepackage{array}
\newcolumntype{?}{!{\vrule width 0.7pt}}

\usepackage[accsupp]{axessibility}  

\usepackage[colorlinks]{hyperref}

\usepackage{orcidlink}

\begin{document}

\title{\texorpdfstring{TRAM: Global Trajectory and Motion of \\ 3D Humans from in-the-wild Videos}
{TRAM: Global Trajectory and Motion of 3D Humans from in-the-wild Videos}
}

\titlerunning{TRAM: Trajectory and Motion of 3D Humans}

\author{Yufu Wang\inst{1}\orcidlink{0000-0001-9907-8382} \and
Ziyun Wang\inst{1}\orcidlink{0000-0002-9803-7949} \and
Lingjie Liu\inst{1}\orcidlink{0000-0003-4301-1474} \and
Kostas Daniilidis\inst{1,2}\orcidlink{0000-0003-0498-0758}}

\authorrunning{Y.~Wang et al.}

\institute{University of Pennsylvania \and Archimedes, Athena RC}

\maketitle

\begin{abstract}
\vspace{-5mm}
We propose TRAM, a two-stage method to reconstruct a human's global trajectory and motion from in-the-wild videos. TRAM robustifies SLAM to recover the camera motion in the presence of dynamic humans and uses the scene background to derive the motion scale. Using the recovered camera as a metric-scale reference frame, we introduce a video transformer model (VIMO) to regress the kinematic body motion of a human. By composing the two motions, we achieve accurate recovery of 3D humans in the world space, reducing global motion errors by a large margin from prior work. 
\url{https://yufu-wang.github.io/tram4d/}

\end{abstract}
\begin{figure*}[h!]
\centering
    \vspace{-8mm}
   \includegraphics[width=1.0\textwidth]{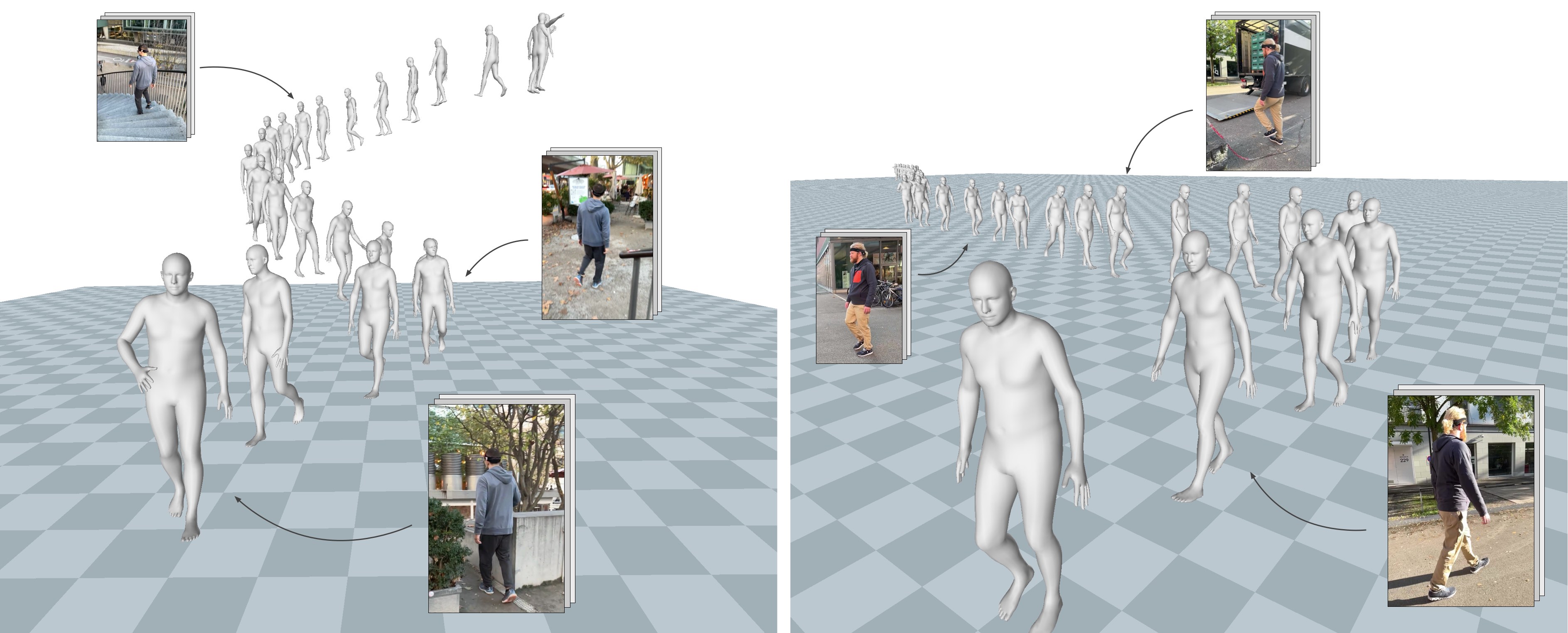}
    \caption{\textbf{Overview.} Given an in-the-wild video, TRAM reconstructs the complete 3D human motion: \textbf{global trajectory and local body motion}, in diverse and long-range scenarios.}
    \label{fig:intro}
    \vspace{-8mm}
\end{figure*}

\section{Introduction}
\label{sec:intro}

Understanding human movement from visual data is a fundamental problem in computer vision. To truly see how we move, we need to capture not only the kinematic body poses but also our global trajectory in the world space as in Fig.\ref{fig:intro}. This task is challenging when the input is a casual video where the human is observed from a moving camera. Without reasoning about the camera motion, most prior studies reconstruct humans in the camera frame, making it difficult to contextualize a human's action in the environment. The ability to capture this complete motion from videos will unlock many applications, such as reasoning about our interaction with the scene~\cite{copilot, ego_body}, imitation learning of life-like behaviors~\cite{sfv, trace&place, martial_art}, gait analysis with metric-scale measurement for health care~\cite{osso, health_survey}, and beyond.

Our observation is that if we can precisely locate the camera trajectory \textbf{in the world frame}, and estimate a person's body motion \textbf{in the camera frame}, then we can recover the person's complete motion in the world by composing the two. This two-stage approach naturally disentangles the motion estimation problem into two well-defined subproblems. The goal of this paper is to  address the two subproblems so that when combined they can capture accurate human motion in the world from a video.

For camera trajectory estimation, Simultaneous Localization and Mapping (SLAM)~\cite{slam_survey} has been widely used in Robotics. However, traditional SLAM faces two key challenges in our context. Firstly, SLAM assumes a static environment while our videos contain moving humans, which decreases the estimation accuracy. Secondly, monocular SLAM only recovers camera trajectory up to an arbitrary scale. To represent camera movement in the metric world frame, we need to estimate a scaling factor. Recent studies propose to infer the scale of camera motion from the observed human motion~\cite{bodyslam, slahmr}. Specifically, after SLAM, these methods jointly optimize human poses and the camera scale, so that the human displacements match a learned locomotion model. However, because the motion models are learned from studio MoCap data, the predicted displacement does not generalize to real-world complexity. Consequently, they struggle to recover complex and long-range trajectories.

In this work, we propose to robustify and metrify SLAM's camera estimation without relying on a human motion model. To make it robust to dynamic humans, we use a masking procedure to dually remove the dynamic regions in the input images and the dense bundle adjustment steps. Enforcing SLAM to only use the background for camera estimation from the beginning reduces the chance of catastrophic failure. To convert the camera estimation to metric scale, we utilize semantic cues from the background. Our insight is that the environment provides rich and reliable information about the scale of the scene. When we watch Spider-Man swing across skyscrapers, we understand the distance he travels despite the lack of locomotion: the environment serves as a scale. We demonstrate that this information can be derived, in a reliable manner, from noisy depth predictions~\cite{zoedepth}. As a result, we recover accurate and metric-scale camera motion to serve as a reference frame for the local human motion. 

The second challenge is the recovery of the kinematic body motion in the camera frame. While seeing great progress from deep regression networks~\cite{spin, hmr}, the de facto per-frame prediction lacks temporal coherence. However, video-based models fall behind in accuracy compared to per-frame models. The main bottleneck in developing a scalable video model is the training cost with videos (concerning larger models) and the lack of video data (compared to the images). To address this issue, our approach leverages a large pre-trained model that has learned rich representations of the human body, and finetunes it on video data. 

We propose Video Transformer for Human Motion (VIMO), that builds on top of the large pre-trained HMR2.0~\cite{hmr2}. VIMO adds two temporal transformers, one to propagate temporal information in the image domain, and the other to propagate information in the motion domain. We keep the ViT-Huge backbone frozen to preserve the learned representations, and finetune the two transformers and the original transformer decoder on videos. Without bells and whistles, this fully ``transformerized'' design achieves state-of-the-art reconstruction accuracy. 

Our method takes a scene-centric point of view in estimating human movements, versus recent methods with a human-centric point of view using motion priors. By solely using the scene background to estimate metric scale camera motion, and always reconstructing humans in the camera frame, our two-stage approach recovers the global trajectory and motion (TRAM) of 3D humans from videos, with a large error reduction in global trajectory from the best published results.

\textbf{Contributions.} (i) We propose a general method TRAM, that recovers human trajectory and motion from in-the-wild videos with a large improvement over prior works. (ii) We demonstrate that the human trajectory can be inferred from SLAM, and provide technical solutions to make monocular SLAM robust and metric-scale in the presence of dynamic humans. (iii) We propose the video transformer model VIMO that builds on top of a large pre-trained image-based model, and demonstrate that this scalable design achieves state-of-the-art pose and shape reconstruction.

\section{Related Work}
\label{sec:related} 

\noindent
\textbf{3D Human Recovery from Images and Videos.} Reconstructing 3D human is most widely formulated as recovering the parameters of a parametric human model, such as SMPL~\cite{smpl} and its followups~\cite{smplx, ghum, mano, frank}. Without reasoning about the camera movement, the majority of the works consider reconstruction in a local coordinate system, often the camera coordinate frame or a multi-view reference frame~\cite{hmr, spin, romp, opt_mosh, multipeople, prob_ego, bev}.

In the optimization paradigm, the SMPL model can be fitted through energy minimization to images~\cite{opt_smplify, opt_silhouettes, opt_multiview, easymocap}, videos~\cite{opt_video, opt_video2, total_capture}, and other sensing modalities~\cite{opt_sensor, 3dpw}. In contrast, regression approaches learn pose and shape representation from large datasets~\cite{h36m, mpii3d, coco}, and predict the parameters given a single image or video frames~\cite{hmr, neuralbodyfit, vibe}. The continuous improvement of accuracy stems, partly from the improvement of datasets~\cite{3dpw, eft, bedlam}, and partly from ingenious designs that imbue domain knowledge into neural architectures~\cite{graphcmr, fast_metro, I2l, pose2mesh, meshform, meshgraphform}. Notable designs includes reprojection alignment~\cite{pymaf, cliff}, part-based reasoning~\cite{pare, hybrik, made}, temporal pooling~\cite{hmmr, tcmr, mpsnet, meva}, and learning-based optimization~\cite{learn2fit, learn2fit2, learn2fit3, refit}. 

In contrast, HMR2.0~\cite{hmr2} demonstrates that a general-purpose transformer architecture~\cite{transformer, vit}, with enough capacity and data, can learn robust representations that in many cases outperform domain-specific designs. In this work, we turn HMR2.0 into a video model with the same principle, by adding two transformer encoders to propagate temporal information from nearby frames. We demonstrate that this simple and fully ``transformerized'' design outperforms existing video models for 3D human regression.

\noindent
\textbf{3D Human Recovery in the World Space.} Reconstructing a person's complete movement through space (global trajectory and motion) is crucial for understanding our actions in the world. Integrating additional sensors or cameras allows methods to capture global human motion reliably~\cite{3dpw, emdb, hps, bodyslam++, 3d_traj, smartmocap}.

This recovery is difficult with in-the-wild videos, where the estimation is with respect to a moving camera. Methods such as GLAMR~\cite{glamr} and D\&D~\cite{d&d} use a human's locomotion to estimate the global trajectory but are not reliable under real-world complexity. In contrast, \cite{egobody, tvshow} densely reconstruct the environment with SfM~\cite{colmap}, and contextualize the human placement with human-scene constraints. This approach requires a priori reconstruction of the environment, which is not always possible with in-the-wild videos. TRACE~\cite{trace} and WHAM~\cite{wham} propose to regress the per-frame pose and translation directly. They achieve great results by learning to regress trajectories from MoCap data, but its reliance on this data as a prior hinders its ability to predict novel trajectories. 

A line of works employs a hybrid approach~\cite{slahmr, pace, bodyslam}. They utilize SLAM~\cite{droid} for camera trajectory estimation (up to a scale) to initialize the human placement and follow with optimization to determine the human poses and the movement scale. They leverage motion models~\cite{humor, nemf} to derive motion scale, but struggle with complex real-world scenarios like navigating stairs or parkour, which are not adequately captured by MoCap data~\cite{amass}. Wang et al.~\cite{ego_global, ego_scene} use a checkerboard to calibrate the motion scale of egocentric videos, but this technique requires objects of exact dimensions to be present.

Our method, TRAM, also recovers camera motion with SLAM, but we differ by deriving the movement scale from the background. By avoiding human motion priors for trajectory estimation, it achieves better generalization to complex scenarios. Notably, we also address SLAM's robustness with dynamic humans.

\section{Method}

Our goal is to recover the complete 3D human motion from videos in the wild. We decompose this motion into its $SE(3)$ root trajectory $\{{\bf{H}}_t\}^T_{t=0}$ in the world frame and the kinematic body motion $\{{\bf{\Theta}}_t\}^T_{t=0}$ represented by a sequence of SMPL poses in the camera frame.

To infer the human trajectory in the world, our approach is to estimate the camera trajectory $\{{\bf{G}}_t\}^T_{t=0}$ and the human's positions with respect to the camera $\{{\bf{T}}_t\}^T_{t=0}$ at each time step. We demonstrate that by making the camera trajectory estimation robust (sec.~\ref{sec:3_2}) and metric-scale (sec.~\ref{sec:3_3}), the human trajectory can be accurately recovered as $\{{\bf{H}}_t\}^T_{t=0} = \{{\bf{G}}_t \circ {\bf{T}}_t\}^T_{t=0}$. 

To reconstruct the kinematic body motions, we propose VIMO (sec.~\ref{sec:3_4}), a video-inference transformer model that reconstructs body motion $\{{\bf{\Theta}}_t\}^T_{t=0}$ and relative positions $\{{\bf{T}}_t\}^T_{t=0}$ in the camera frame. 

\subsection{Preliminary: 3D Human Model}
We use SMPL~\cite{smpl} to represent the 3D human body. The SMPL model is a parametric mesh model $\mathcal{M}(\theta, \beta, r, \pi) \in \mathbb{R}^{6890\times3}$, where $\theta \in \mathbb{R}^{23\times3}$ are the relative rotations of the 23 body joints, $\beta \in \mathbb{R}^{10}$ is the shape parameter, and $r \in \mathbb{R}^3$ and $\pi \in \mathbb{R}^{3}$ are the root orientation and translation w.r.t the camera. In our context of motion in the world frame, ${\bf{T}}_t = \{r_t, \pi_t\}$ represents the relative position, and ${\bf{\Theta}}_t = \{\theta_t, \beta_t\}$ represents the local body pose and shape. The scale of the SMPL mesh is in metric units, representing the real size of the human in meters.

\begin{figure*}[t!]
\centering
   \includegraphics[width=1.0\textwidth]{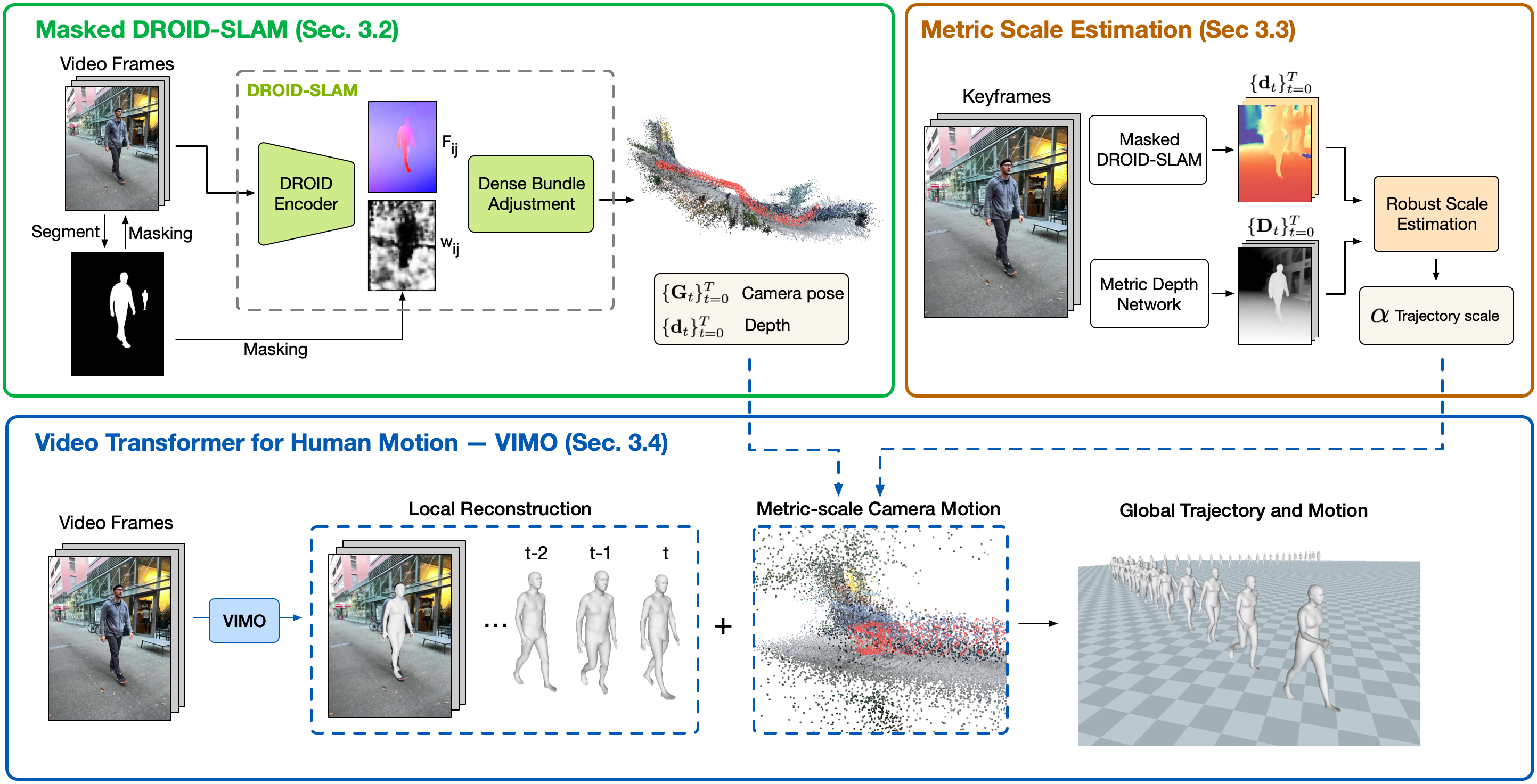}
    \caption{\textbf{Overview of TRAM.} Top-left: given a video, we first recover the relative camera motion and scene depth with DROID-SLAM, which we robustify with dual masking (Sec.~\ref{sec:3_2}). Top-right: we align the recovered depth to metric depth prediction with an optimization procedure to estimate metric scaling (Sec.~\ref{sec:3_3}). Bottom: We introduce VIMO to reconstruct the 3D human in the camera coordinate (Sec.~\ref{sec:3_4}), and use the metric-scale camera to convert the human trajectory and body motion to the global coordinate. 
    }
    \label{fig:slam_method}
\end{figure*}

\subsection{Masked DROID-SLAM}
\label{sec:3_2}

We utilize DROID-SLAM~\cite{droid} to recover the camera trajectory $\{{\bf{G}}_t\}^T_{t=0}$ from monocular videos. In this paper, we propose to reduce the negative effect of dynamic objects through a two-step masking.

For a input video frame ${\bf{I}}_i$, DROID first computes the 2D flows ${\bf{F}}_{ij} \in {\mathbb R}^{h\times w \times2}$ and its confidence ${\bf{w}}_{ij} \in {\mathbb R}^{h\times w \times2}$ with respect to nearby keyframes $\{ {\bf{I}}_j\}$. Then the dense bundle adjustment layer (DBA) optimizes the relative camera pose ${\bf{G}}_{ij} \in SE(3)$ and depth of the current frame ${\bf{d}}_i \in {\mathbb R}^{h \times w}$ by solving the following objective with Gauss-Newton.
\begin{equation}
    E(G, d) = \sum_{(i,j)} \parallel p_{ij} - \Pi(G_{ij} \circ \Pi^{-1}(p_i, d_i)) \parallel^2_{\scriptscriptstyle \sum_{ij}} 
    {\scriptstyle \sum_{ij} = \text{diag} \textbf{w}_{ij}}
    \label{eq:bundle_adjustment}
\end{equation}
where ${\bf{p}}_i$ are the pixel coordinates and ${\bf{p}}_{ij} = {\bf{p}}_{i} + {\bf{F}}_{ij}$. This objective minimizes the flow reprojection error weighted by confidence. Global bundle adjustment and loop-closure are applied after all frames are processed. 

The predicted confidence ${\bf{w}}_{ij}$ allows DROID to down-weight correspondences with high uncertainty from the DBA process, making it robust to small moving objects. However, when the dynamic object occupies a larger area, the predicted confidence may not be accurate. Pixel coordinates that violate the static scene assumption degrade the accuracy of camera motion estimation. 

We propose to dually mask the input image ${\bf{\hat I}}_i = \text{mask}({\bf{I}}_i)$ and the predicted confidence ${\bf{\hat w}}_{ij} = \text{mask}({\bf{w}}_{ij})$, setting the dynamic region to value zero as shown in Fig.\ref{fig:slam_method}. Masking the flow confidence ${\bf{w}}_{ij}$ is equivalent to removing the dynamic object coordinates from the calculation of reprojection error in Eq.\ref{eq:bundle_adjustment}. This step ensures the DBA only uses background pixels to estimate camera motion, making it robust to large moving objects. Additionally, we also find it beneficial to mask the input images. DROID's learned encoder uses both local and global features to predict dense flow on the image plane. The global feature provides useful context for flow estimation, but large moving objects may contribute motion cues that negatively impact other regions. Masking the input image helps to mitigate this effect on the global feature. We can predict accurate masks of dynamic objects using an object detector~\cite{yolo} and Segment Anything~\cite{sam}. In this paper, we focus on humans, but other categories can also be detected and masked out. 

Our experiments show that the two masking steps are essential, preventing DROID-SLAM from diverging in many sequences where moving humans occupy a large foreground. Masked DROID allows us to recover accurate camera motion, from which we can then infer the human's trajectory as seen from the camera.

\subsection{Trajectory Scale Estimation} 
\label{sec:3_3}
Masked DROID can recover camera trajectory and scene structure up to an arbitrary scale, but we need them to be in metric scale to represent motion in the world frame and be compatible with SMPL. We show that a scaling parameter that indicates the actual size of the scene can be solved from the semantic of the scene. 

Common objects such as buildings, cars, and trees are of known sizes, and as humans we have a mental model that reasons about their spatial scale and relationship. This spatial reasoning capacity is partially presented in a depth prediction network. We can use metric depth prediction to reason about the scale of the camera motion and the scene reconstructed by SLAM. In this paper, we use ZoeDepth~\cite{zoedepth} to predict metric depth for video frames. Given a keyframe ${\bf{I}}_{i}$, ZoeDepth predicts the scene depth ${\bf{D}}_{i} \in {\mathbb R}^{h\times w}$ in meters. DROID returns the depth ${\bf{d}}_{i}$ in a random unit. By solving for a scaling term $\alpha$ that aligns $\alpha * {\bf{d}}_{i}$ to ${\bf{D}}_{i}$, we can re-scale the reconstructed scene and the camera trajectory to the metric unit. 

However, depth prediction is usually noisy. The network can under or over-estimate the depth in certain regions of the image. For example, we find ZoeDepth to often underestimate the distance in the sky region. Similarly, the depth predictions are not temporally consistent and can be inaccurate in many frames. Therefore, we need to derive the scale robustly. To mitigate the effect of bad depth prediction at certain frames, we will solve for the scale for each keyframe independently and take the median over all the keyframes. To solve for the scale for each keyframe, we minimize the energy with robust least square as
\begin{equation}
    E(\alpha) = \sum_{(h,w)} \rho(\alpha * {\bf{d}}_{i} - {\bf{D}}_{i})
    \label{eq:least_square}
\end{equation}
where $\rho$ is the German-McClure robust loss function with the summation over the entire image. We use BFGS for the optimization. Furthermore, we find that depth predictions are less accurate in the far region such as the sky or building in the far distance. Therefore, we set thresholds to exclude the far region from the optimization, essentially using only the middle region where depth prediction is more reliable to solve for the scale. 

The thresholding and robust least squares handle noisy areas in a frame, and the median allows us to use the statistics of the whole trajectory. Overall, our experiments show that this procedure provides a good estimate for the scale of the scene, and is more reliable than inferring scale from human motion models.

\subsection{Video Transformer for Human Motion}
\label{sec:3_4}

Human pose and shape regression networks can now reconstruct 3D people from a single image in the general case, but videos provide temporal constraints to help reconstruct motions that look natural and smooth. In this work, we propose {\bf VI}deo Transformer for Human {\bf MO}tion (VIMO) to accomplish this goal. 

A recent success recipe in vision is to finetune a large pre-trained model for downstream tasks. HMR2.0~\cite{hmr2} demonstrates state-of-the-art reconstruction by building on top of a pre-trained ViT-H~\cite{mae, vitpose} and scaling up its training data and compute. Our design philosophy for VIMO is therefore twofold: we aim to utilize the rich knowledge in a large pre-trained model, and make flexible and scalable architecture design whose performance can scale with data. 

\begin{figure*}[t!]
\centering
   \includegraphics[width=0.98\textwidth]{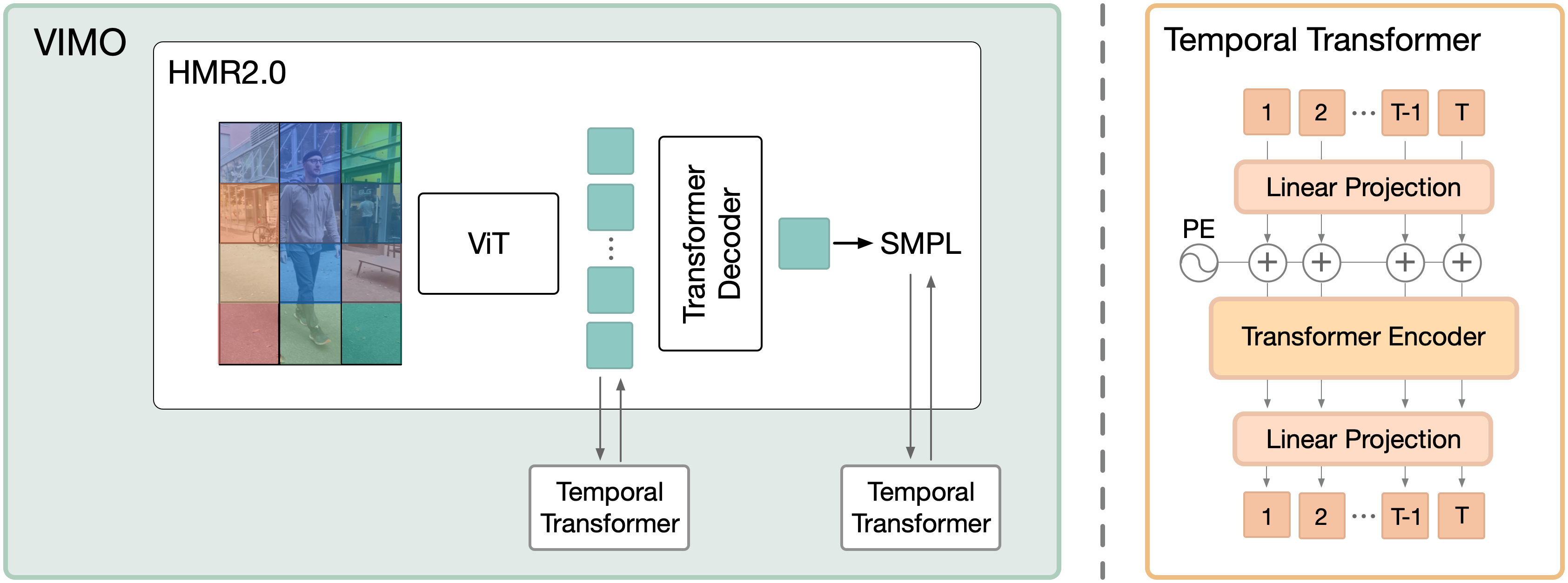}
    \caption{\textbf{Video transformer VIMO} builds on top of the large pre-trained HMR2.0 and adds two temporal transformers to propagate information across video frames. {\color{orange!90}Right}: the temporal transformers use the same encoder-only architecture. \raisebox{0.6mm}{\fcolorbox{black}{orange!50}{\rule{1pt}{0pt}\rule{0pt}{1pt}}} represents patch tokens at the same spatial location across time in the first temporal module, and represents SMPL poses across time in the second temporal module. More details are included in the supplementary.}
    \label{fig:vimo}
\end{figure*}

Toward this goal, we turn the image-based HMR2.0 into a video model by adding two temporal transformers, as shown in Fig. \ref{fig:vimo}. HMR2.0 uses a ViT-H to extract image features and a standard transformer decoder with a zero token to cross-attend the image features and regress the final outputs. The two new temporal transformers from VIMO use the encoder-only architecture but distinctively propagate temporal information in the image domain and the human motion domain. 

The first transformer applies attention across time on each patch token from ViT. This operation is repeated at different spatial locations independently. The combination of ViT and this temporal transformer can be considered as a factorized spatial-temporal model~\cite{vivit}, where the ViT applies attention across space and the temporal transformer applies attention across time. The role of this layer is to use temporal correlation, such as appearance and motion cues, to make the image features more accurate and robust. 

The second transformer encodes and decodes a sequence of SMPL poses. The goal here is to learn a prior on the human motion space so that noisy poses can be corrected to become smooth motion. Previous work largely learns temporal models in the global feature space~\cite{vibe, thmmr, meva}. We conjecture that the latent space before the regressor is sub-optimal to learn a motion model, as this space entangles many other information such as shape, camera, and image features. In contrast, studies in motion generation and denoising~\cite{actor, posebert} show that motion, represented as a sequence of poses, can be directly encoded and decoded with transformers. Therefore, we apply this general architecture on SMPL poses $\{\theta_t, r_t\}$ directly. While such a motion model can be pre-trained with MoCap data, we demonstrate end-to-end learning from videos. 

We keep the ViT backbone frozen and finetune VIMO on video data with the following losses.
\begin{equation}
\mathcal{L} = \lambda_{2D}\mathcal{L}_{2D} + \lambda_{3D}\mathcal{L}_{3D} + \lambda_{SMPL}\mathcal{L}_{SMPL}+ \lambda_{V}\mathcal{L}_{V}
\label{eq:loss}
\end{equation}
Each term is calculated as 
\begin{align*}
\mathcal{L}_{2D} &= ||\mathcal{\hat{J}}_{2D} - \Pi(\mathcal{J}_{3D})||^2_F \\
\mathcal{L}_{3D} &= ||\mathcal{\hat{J}}_{3D} - \mathcal{J}_{3D}||^2_F \\
\mathcal{L}_{SMPL} &= ||\hat{\Theta} - \Theta||^2_2 \\
\mathcal{L}_{V} &= ||\hat{V} - V||^2_F
\label{eq:terms}
\end{align*}
where $\mathcal{J}_{3D}$ and $V$ are the 3D joints and vertices obtained from the SMPL model, and the hat operator denotes the ground truth of that variable. Our experiments show that both temporal transformers are essential for VIMO to reconstruct accurate and smooth motion from videos.

\section{Experiments}

\textbf{Datasets.} We use three datasets to train our video model VIMO: 3DPW~\cite{3dpw}, Human3.6M~\cite{h36m}, and BEDLAM~\cite{bedlam}. As models are trained on increasingly diverse data, we create a baseline by finetuning HMR2.0 on the same three datasets as a fair comparison to validate our design. We evaluate human motion and shape reconstruction on 3DPW and EMDB (subset 1)~\cite{emdb}. For human trajectory recovery, we evaluate on the EMDB dataset (subset 2), which contains 25 sequences with ground truth for both human and camera trajectory. 

\noindent
\textbf{Implementation.} We train VIMO for 100K iterations using AdamW~\cite{adamw} with weight decay of 0.01 and a batch size of 24 sequences, each with a 16-frame window. We use a learning rate of 1e-5 for the transformer decoder and 3e-5 for the two temporal transformers.  We include more details about the architecture in the supplementary. For masking, we use detections from YOLOv7~\cite{yolo} as the prompt for the Segment Anything Model~\cite{sam}. We use ZoeDepth~\cite{zoedepth} to predict metric depth from videos. 

\noindent
\textbf{Evaluation metrics.} For pose and shape reconstruction, we use the common reconstruction metrics: mean per-joint error (MPJPE), Procrustes-aligned per-joint error (PA-MPJPE), and per-vertex error (PVE). To evaluate the motion smoothness, we compute acceleration error (ACCEL) against the ground truth acceleration. 

For human trajectory evaluation, we slice a sequence into 100-frame segments and evaluate 3D joint error after aligning the first two frames (W-MPJPE$_{100}$) or the entire segment (WA-MPJPE$_{100}$)~\cite{slahmr}. Following WHAM~\cite{wham}, we evaluate root translation error normalized by the total displacement (RTE in $\%$) after rigid alignment without scaling. In addition, we compute egocentric-frame root velocity error (ERVE) to measure the root motion accuracy.

For camera trajectory evaluation, we follow SLAM literature and evaluate absolute trajectory error (ATE)~\cite{ate}, which performs rigid alignment with scaling to align the camera trajectory with ground truth before computing error. To evaluate the accuracy of our scale estimation, we also evaluate ATE using our estimated scale (ATE-S)~\cite{pace}.

\subsection{Comparison on Camera Trajectory Recovery}
\textbf{Dynamic masking}. We first validate the two-step masking procedure by evaluating the camera trajectory accuracy on EMDB. We create two baselines with ORB-SLAM2~\cite{orb2} and DROID-SLAM~\cite{droid}. 

As shown in Table \ref{tab:cam}, the default DROID-SLAM has a large error on EMDB due to dynamic humans. Compared to DROID, ORB-SLAM2 is less affected by dynamic humans, likely due to its outlier rejection procedure (e.g. RANSAC) and only using sparse points. However, ORB-SLAM2 loses track in 9 out of 25 sequences due to the sudden loss of feature points. Masking improves the results of both methods, particularly in longer sequences. 

\begin{figure*}[t!]
\vspace{-3mm}
\centering
   \includegraphics[width=1.0\textwidth]{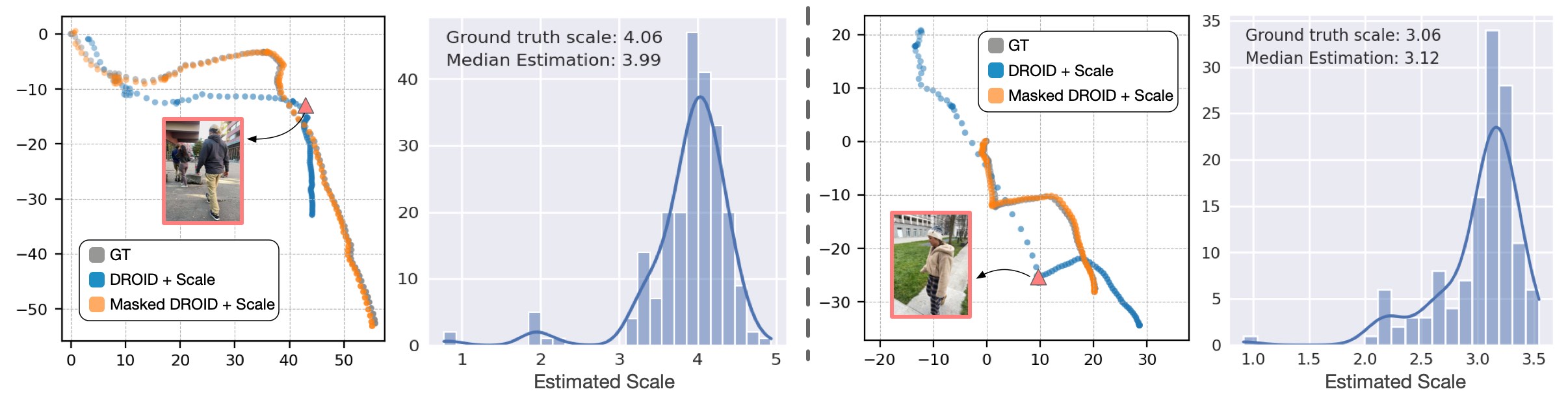}
   \vspace{-6mm}
    \caption{\textbf{Camera trajectory estimation.} With dynamic humans in the scene, the default DROID-SLAM tends to diverge. The two-step masking makes it robust. In addition, our procedure estimates a reasonable metric scale for the cameras. }
    \label{fig:cam_traj}
\end{figure*}

Masking humans in the input images improves the accuracy. The improvement is more significant with masking in the dense bundle adjustment process. By masking both the input and the DBA, we achieve the largest improvement over the baseline. We observe that all previously diverged sequences now have a more reasonable trajectory, as shown in Figure~\ref{fig:cam_traj}.

\begin{table*}[t!]
\centering
\setlength{\tabcolsep}{3pt}
\resizebox{0.90\textwidth}{!}
{\small{
\begin{tabular}{l?cccc}
\cmidrule{1-5}
& \multicolumn{4}{c}{EMDB 2 (ATE)} \\
\cmidrule(lr){2-5}
Methods & Short(5) & Medium(10) & Long(10) & Average \\
\cmidrule{1-5}
ORB-SLAM2  & 0.08 & 0.29 & 2.08 & 1.05 \\
ORB-SLAM2 + Mask Image & 0.38 & 0.60 & 1.10 & 0.79 \\
\cmidrule{1-5}
DROID-SLAM & 0.40 & 2.55 & 3.31 & 2.42 \\
DROID-SLAM + Mask Image  & 0.36 & 0.63 & 2.74 & 1.42\\
DROID-SLAM + Mask DBA  & 0.45 & 0.42 & 1.63 & 0.91\\
\textbf{Masked DROID} & \textbf{0.32} & \textbf{0.20} & \textbf{0.44} & \textbf{0.32} \\
\cmidrule{1-5}
\end{tabular}
}}
\caption{Evaluation of camera estimation \textbf{with ground truth scale (ATE)}. Results are grouped according to sequence length: short(<20m), medium(<60m) and long(>60m). Parenthesis denote the number of sequences. ORB-SLAM2 fails in 9/25 sequences so its results are calculated with the other 16 sequences. ATE is in $m$.
}
\vspace{-2mm}
\label{tab:cam}
\end{table*}

\begin{table*}[t!]
\vspace{-3mm}
\centering
\setlength{\tabcolsep}{3pt}
\resizebox{0.90\textwidth}{!}
{\small{
\begin{tabular}{l?cccc}
\cmidrule{1-5}
& \multicolumn{4}{c}{EMDB 2 (ATE-S)} \\
\cmidrule(lr){2-5}

Methods & Short(5) & Medium(10) & Long(10) & Average \\
\cmidrule{1-5}
Masked DROID + ZoeDepth & \textbf{0.33} & 2.29 & 5.27 & 3.09 \\
\textbf{Masked DROID + Scale est.} & 0.48 & \textbf{0.62} & \textbf{0.78} & \textbf{0.66} \\
\cmidrule{1-5}
\end{tabular}
}}
\caption{Evaluation of camera estimation \textbf{with estimated scale (ATE-S)}. Naively using ZoeDepth predictions as depth input for DROID results in large error. The proposed method produces good scale estimation. ATE-S is in $m$. 
}
\vspace{-2mm}
\label{tab:cam_metric}
\end{table*}

\textbf{Metric scaling}. The above evaluation calculates ATE with the ground truth scale, but our target application requires us to estimate the scale. We evaluate this error in Table~\ref{tab:cam_metric}. As a baseline, we give ZoeDepth prediction directly to DROID using its RGB-D mode. This naive combination causes large errors because incorrect depth predictions lead the bundle adjustment to wrong solutions. Currently, metric depth prediction is not accurate enough to replace RGB-D inputs. Our method circumvents noisy depth prediction with robust optimization and uses the median of the whole trajectory. Compared to using the ground truth scale, it increases the error only by 30cm, a gap that will likely become smaller with continuous improvement of depth prediction~\cite{depthanything, marigold}. 

We visualize the distribution of the estimated scales among keyframes in a sequence in Figure \ref{fig:cam_traj}. As shown, not all frames are suitable for estimating scale due to noisy or incorrect depth prediction. Depth prediction can also incur a large bias in frames where there is a lack of references. Using the median of the whole sequence approximates the ground truth scale well. Figure \ref{fig:cam_traj} shows two sequences in which DROID diverges without masking. Our method estimates accurate and metric-scale camera trajectory, which becomes a reference frame when reasoning about the human motion in the world.

\begin{table*}[t!]
\centering
\setlength{\tabcolsep}{3pt}
\renewcommand{\arraystretch}{1.1}
\resizebox{0.99\textwidth}{!}
{\small{
\begin{tabular}{l?cccccc}
\cmidrule{1-6}
& \multicolumn{5}{c}{EMDB 2} \\
\cmidrule(lr){2-6}

Models & PA–MPJPE & WA-MPJPE$_{100}$ & W-MPJPE$_{100}$ & RTE & ERVE \\
\cmidrule{1-6}
TRACE~\cite{trace} & 58.0 & 529.0 & 1702.3 & 17.7 & 370.7 \\
GLAMR~\cite{glamr}  & 56.0 & 280.8 & 726.6 & 11.4 & 18.0 \\
SLAHMR~\cite{slahmr} & 61.5 & 326.9 & 776.1 & 10.2 & 19.7 \\
WHAM (w/ DROID )~\cite{wham} & 38.2 & 133.3 & 343.9 & 4.6 & 14.7 \\
\cmidrule{1-6}
\textbf{TRAM} & \textbf{38.1} & \textbf{76.4} & \textbf{222.4} & \textbf{1.4} & \textbf{10.3} \\
\cmidrule{1-6}
\end{tabular}
}}
\caption{\textbf{Evaluation of human global trajectory and motions}. RTE is in $\%$, ERVE is in $mm/\text{frame}$, and the other pose metrics are in $mm$. Ordered by RTE.
}
\vspace{-2mm}
\label{tab:human_traj}
\end{table*}

\begin{figure*}[t!]
\vspace{-2mm}
\centering
   \includegraphics[width=0.98\textwidth]{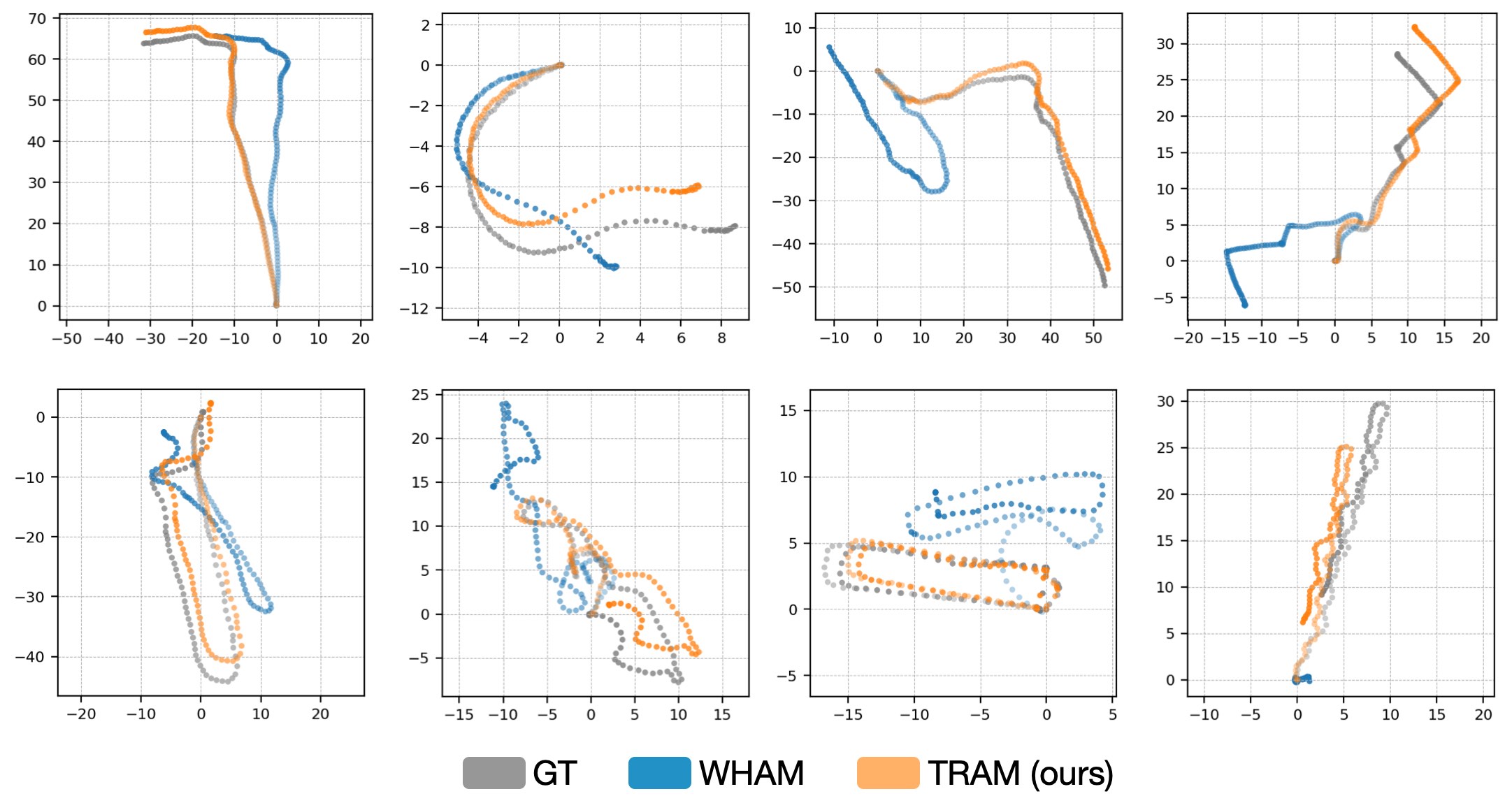}
   \vspace{-1mm}
    \caption{\textbf{Human global trajectory} on EMDB. Compared to WHAM, our method produces less drift and a more accurate scale for complex and long-range tracks.}
    \label{fig:human_traj}
\end{figure*}

\subsection{Comparison on Human Trajectory Recovery}
With accurate and metric-scale camera trajectory, we can infer the human trajectory as proposed in Section 3, using VIMO to reconstruct the 3D human with respect to the camera frame. As shown in Table \ref{tab:human_traj}, we improve global pose and trajectory accuracy by a large margin. Particularly, we achieve a 60\% error reduction in the root trajectory estimation (RTE). This confirms our observation that accurate camera recovery is essential in estimating human motion in the world frame. 

We visualize the human trajectory in the xy-plane in Figure \ref{fig:human_traj}. As observed, WHAM can recover good trajectories when the humans walk in mostly straight lines. If the trajectories involve large curves, WHAM's learned regressor cannot recover an accurate trajectory. In contrast, we recover complex trajectories with a minimum drift and a more accurate scale. Moreover, WHAM fails when the motion is outside the MoCap data, such as walking down stairs or riding a skateboard, as shown in Figure \ref{fig:human_motion}. We can recover such complex trajectories because our method does not depend on a learned prior from MoCap data.

\subsection{Comparison on Human Body Motion Reconstruction}

\begin{figure*}[h!]
\centering
   \includegraphics[width=0.98\textwidth]{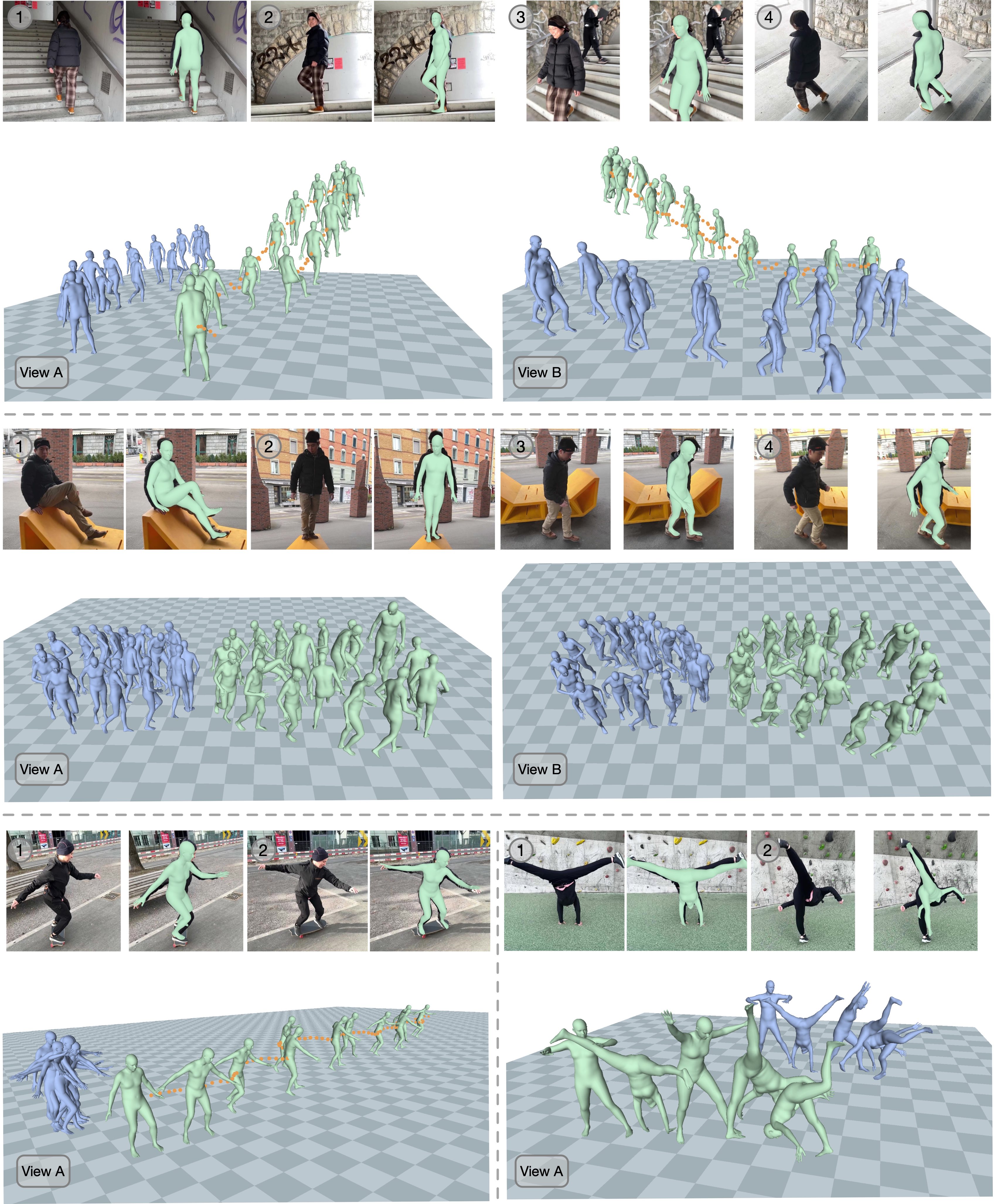}
   \vspace{-1mm}
    \caption{\textbf{Human motion and trajectory:} WHAM \raisebox{1mm}{\fcolorbox{black}{blue!50}{\rule{1pt}{0pt}\rule{0pt}{1pt}}} vs Ours \raisebox{1.mm}{\fcolorbox{black}{green!30}{\rule{1pt}{0pt}\rule{0pt}{1pt}}}. We produce more accurate motion and tracks that generalize to diverse terrain and motion complexity. WHAM's results are from the version that uses ground truth gyro.}
    \label{fig:human_motion}
\end{figure*}

\begin{table*}[t!]
\centering
\setlength{\tabcolsep}{3pt}
\renewcommand{\arraystretch}{1.1}
\resizebox{0.99\textwidth}{!}
{\small{
\begin{tabular}{cl?cccc?cccc}
\cmidrule[0.75pt]{1-10}
& & \multicolumn{4}{c}{3DPW (14)} & \multicolumn{4}{c}{EMDB (24)} \\
\cmidrule(lr){3-6} \cmidrule(lr){7-10}

& Models & \scriptsize{PA-MPJPE} & \scriptsize{MPJPE} & \scriptsize{PVE} & \scriptsize{Accel} & \scriptsize{PA-MPJPE} & \scriptsize{MPJPE} & \scriptsize{PVE} & \scriptsize{Accel} \\
\cmidrule{1-10}

\multirow{6}{1em}{\rotatebox[origin=c]{90}{per-frame}} 

& SPIN~\cite{spin}  &  59.2 & 96.9 & 112.8 & 31.4 & 87.1 & 140.7 & 166.1 & 41.3 \\

& PARE~\cite{pare}  & 46.5 & 74.5 & 88.6 & -- & 72.2 & 113.9 & 133.2 & -- \\

& CLIFF~\cite{cliff} & 43.0 & 69.0 & 81.2 & 22.5 & 68.3 & 103.3 & 123.7 & 24.5 \\

& HybrIK~\cite{hybrik}  & 41.8 & 71.6 &  82.3 & -- & 65.6 & 103.0 & 122.2 & -- \\

& HMR2.0~\cite{hmr2}  & 44.4 & 69.8 & 82.2 & 18.1 & 60.7 & 98.3 & 120.8 & 19.9 \\

& ReFit~\cite{refit}  & 40.5 & 65.3 & 75.1 & 18.5 & 58.6 & 88.0 & 104.5 & 20.7 \\
\cmidrule{1-10}

\multirow{7}{1em}{\rotatebox[origin=c]{90}{temporal}} 

& TCMR~\cite{tcmr} & 52.7 & 86.5 & 101.4 & 6.0 & 79.8 & 127.7 & 150.2 & 5.3 \\

& VIBE~\cite{vibe}  & 51.9 & 82.9 & 98.4 & 18.5 & 81.6 & 126.1 & 149.9 & 26.5 \\

& MPS-Net~\cite{mpsnet}   & 52.1 & 84.3 & 99.0 & 6.5 & 81.4 & 123.3 & 143.9 & 6.2 \\

& GLoT~\cite{glot}  & 50.6 & 80.7 & 96.4 & 6.0 & 79.1 & 119.9 & 140.8 & 5.4 \\

& GLAMR~\cite{glamr} & 51.1 & -- & -- & 8.0 & 73.8 & 113.8 & 134.9 & 33.0 \\

& TRACE~\cite{trace} & 50.9 & 79.1 & 95.4 & 28.6 & 71.5 & 110.0 & 129.6 & 25.5 \\

& WHAM (ViT)~\cite{wham}  & 35.9 & \textbf{57.8} & \textbf{68.7} & 6.6 & 50.4 & 79.7 & 94.4 & 5.3 \\

\cmidrule[0.5pt]{1-10}

& \textbf{HMR2.0(ft)}  & 37.3 & 63.2 & 74.3 & 14.8 & 49.7 & 82.7 & 95.3 & 20.5 \\
&\textbf{VIMO} & \textbf{35.6} & 59.3 & 69.6 & \textbf{4.9} & \textbf{45.7} & \textbf{74.4} & \textbf{86.6} & \textbf{4.9} \\

\cmidrule[0.75pt]{1-10}
\end{tabular}
}}
\caption{\textbf{Comparison of mesh reconstruction} on the 3DPW and EMDB datasets. HMR2.0(ft) is our baseline by finetuning HMR2.0b on the same training data as VIMO. Parenthesis denotes the number of body joints used to compute errors for the dataset. Bold numbers denote the best performance. Accel is in $m/s^2$, and others are in $mm$. 
}
\vspace{-2mm}
\label{tab:quant_hps}
\end{table*}

We evaluate human mesh reconstruction from VIMO in Table \ref{tab:quant_hps}. Without domain-specific designs, VIMO outperforms all other methods in both reconstruction accuracy and motion smoothness. 

To get a fair comparison, we create a baseline HMR2.0(ft), by finetuning the transformer decoder of HMR2.0 with the same training data and procedure as VIMO. HMR2.0(ft) has a higher benchmark performance than the official release because it uses additional data from 3DPW and BEDLAM. VIMO achieves consistent improvements over the baseline, showing the effectiveness of the two temporal transformers. 

We conduct ablations on VIMO. As shown in Table \ref{tab:vimo_ablation}, both temporal transformers are important. Removing the token temporal transformer decreases the reconstruction accuracy, indicating that there is information gain in the image feature domain by considering neighboring frames with attention. Removing the motion transformer decreases motion smoothness, as it plays a key role in denoising to produce smooth and natural motion. 

We provide more qualitative results in Figure \ref{fig:human_motion}. VIMO's reconstruction aligns well with the input frames and handles complex poses gracefully. We attribute this robustness to its pre-training from HMR2.0. Freezing the ViT-Huge backbone preserves the recognition power. Since the whole VIMO architecture is made up of transformers, this is a general and scalable design that can take advantage of scaling the model size, compute, and data~\cite{scaling_vit, scaling_vit2, train_large}.  

\subsection{Limitations}
While TRAM works well on datasets, we detail the following challenges for future research. Firstly, SLAM's dependence on known focal length limits its applicability in many in-the-wild cases. Future research will need to address the estimation of focal length during bundle adjustment~\cite{scene_coord, flowmap}. Secondly, depth estimation tends to be less accurate with extreme focal length, and methods that consider focal length will be beneficial~\cite{metric3d}. Lastly, our method follows a strict separation of camera and human motion but a joint optimization in the end could be beneficial~\cite{pace}. Joint optimization with proper physics priors will improve implausible movements such as foot sliding and penetration~\cite{multiphys}.

\begin{table*}[t!]
\centering
\setlength{\tabcolsep}{3pt}
\renewcommand{\arraystretch}{1.1}
\resizebox{0.99\textwidth}{!}
{\small{
\begin{tabular}{cl?cccc?cccc}
\cmidrule[0.75pt]{1-10}
& & \multicolumn{4}{c}{3DPW (14)} & \multicolumn{4}{c}{EMDB (24)} \\
\cmidrule(lr){3-6} \cmidrule(lr){7-10}

& Models & \scriptsize{PA-MPJPE} & \scriptsize{MPJPE} & \scriptsize{PVE} & \scriptsize{Accel} & \scriptsize{PA-MPJPE} & \scriptsize{MPJPE} & \scriptsize{PVE} & \scriptsize{Accel} \\
\cmidrule{1-10}

& HMR2.0(ft)  & 37.3 & 63.2 & 74.3 & 14.8 & 49.7 & 82.7 & 95.3 & 20.5 \\
& + Tokens Attention  & 36.3 & 59.5 & \textbf{69.5} & 8.9 & \textbf{45.4} & \textbf{74.1} & \textbf{85.6} & 11.6 \\
& + Motion Attention  & 37.0 & 60.5 & 71.1 & 4.9 & 48.4 & 78.1 & 89.2 & 5.2 \\
&\textbf{VIMO} & \textbf{35.6} & \textbf{59.3} & 69.6 & \textbf{4.9} & 45.7 & 74.4 & 86.6 & \textbf{4.9} \\

\cmidrule[0.75pt]{1-10}
\end{tabular}
}}
\caption{\textbf{Ablation on VIMO}. Removing either temporal transformer decrease reconstruction accuracy or motion smoothness. The proposed VIMO recovers accurate and smooth motion.  
}
\vspace{-5mm}
\label{tab:vimo_ablation}
\end{table*}

\section{Conclusions}
We presented TRAM, a new two-stage method to recover the global human trajectory and body motion from in-the-wild videos with moving cameras. TRAM is efficient and accurate, improving global human reconstruction by large margins. We also introduced VIMO, a video transformer model for regressing the local human body motion. VIMO is simple and scalable, and
outperforms prior models in different pose and shape reconstruction benchmarks. 

\bigbreak
\noindent
\textbf{Acknowledgements} The authors appreciate the support of the following grants: NSF NCS-FO 2124355, NSF FRR 2220868, NSF IIS-RI 2212433.


%
%
\bibliographystyle{splncs04}
\bibliography{main}
\clearpage




\end{document}